\documentclass[journal]{IEEEtran}
\usepackage{amsmath, amssymb, amsfonts, amsthm, latexsym}
\usepackage{multirow}
\usepackage{bm}
\usepackage{url}
\usepackage{enumerate}
\newtheorem{prop}{Proposition}

\usepackage{mathtools}
\mathtoolsset{showonlyrefs=true} 

\usepackage{color}

\usepackage{latexsym}
 
\DeclareUnicodeCharacter{221E}{\ensuremath{\infty}}

\usepackage{cite}

\begin{document}
\title{Model Compression Method for S4 with Diagonal State Space Layers using Balanced Truncation}
\author{Haruka Ezoe and Kazuhiro Sato\thanks{H. Ezoe and K. Sato are with the Department of Mathematical Informatics, Graduate School of Information Science and Technology, The University of Tokyo, Tokyo 113-8656, Japan, email: ezoha7@g.ecc.u-tokyo.ac.jp (H. Ezoe), kazuhiro@mist.i.u-tokyo.ac.jp (K. Sato) }}
\maketitle
\thispagestyle{empty}
\pagestyle{empty}

\begin{abstract}
To implement deep learning models on edge devices, model compression methods have been widely recognized as useful.
However, it remains unclear which model compression methods are effective for Structured State Space Sequence (S4) models incorporating Diagonal State Space (DSS) layers, tailored for processing long-sequence data.
In this paper, we propose to use the balanced truncation, a prevalent model reduction technique in control theory, applied specifically to DSS layers in pre-trained S4 model as a novel model compression method.
Moreover, we propose using the reduced model parameters obtained by the balanced truncation as initial parameters of S4 models with DSS layers during the main training process.
Numerical experiments demonstrate that our trained models combined with the balanced truncation surpass conventionally trained models with Skew-HiPPO initialization in accuracy, even with fewer parameters.
Furthermore, our observations reveal a positive correlation: higher accuracy in the original model consistently leads to increased accuracy in models trained using our model compression method,
suggesting that our approach effectively leverages the strengths of the original model.
\end{abstract}


\IEEEpeerreviewmaketitle

\section{Introduction} \label{sec:intro}
In recent years, deep learning models have garnered substantial attention due to their versatility across a range of applications, including sequence prediction, natural language translation, speech recognition, and audio generation \textcolor{black}{\cite{lim2021time, mehrish2023review, pandey2023natural}}. These models' ability to understand and predict sequential data underpins their success in these domains.
A critical aspect of these models' effectiveness is their capacity to capture dependencies between sequential data points, a fundamental requirement for achieving high levels of performance in tasks involving time series or sequential input.
For instance, the Transformer \cite{vaswani2017attention} is effective in capturing short-range dependencies in sequential data, and achieved a state-of-the-art BLEU score of 41.0 on the WMT 2014 English-to-French translation task. 
However, the Transformer's ability to capture long-range dependencies in time series data is limited, leading to the loss of temporal information due to its permutation-invariant self-attention mechanism \cite{zeng2023transformers}.

Contrary to the limitations observed in the Transformer model, the Structured State Space Sequence (S4) model, as introduced in \cite{gu2022efficiently}, demonstrates exceptional capability in capturing long-range dependencies within sequential data. This effectiveness is largely attributed to the innovative use of HiPPO initialization \cite{gu2020hippo}, a technique specifically designed to enhance model performance by leveraging the principles of the state space model (SSM) from control theory.
Notably, the S4 model has shown to surpass conventional models, including the Transformer, in Long Range Arena (LRA) tasks \cite{tay2021long}, signifying a substantial advancement in handling sequential data. Further refinement of the S4 architecture led to the introduction of the Diagonal State Space (DSS) layers \cite{gupta2022diagonal}, offering a simplified yet effective version of the original S4 model, maintaining its high performance with a more streamlined architecture.
In addition to the original and simplified S4 models, several deep learning models related to the SSM, such as H3 \cite{fu2022hungry}, Hyena \cite{poli2023hyena}, S4D \cite{gu2022parameterization}, S4ND \cite{nguyen2022s4nd},
S5 \cite{smith2023simplified}, SSSD \cite{lopez2023diffusion},
and Mamba \cite{gu2023mamba}, have been proposed.
Generally, in tasks requiring long-range dependency modeling, these deep learning models tend to perform better with a larger number of parameters.

\begin{figure}[t]
    \centering
    \includegraphics[width=8.5cm]{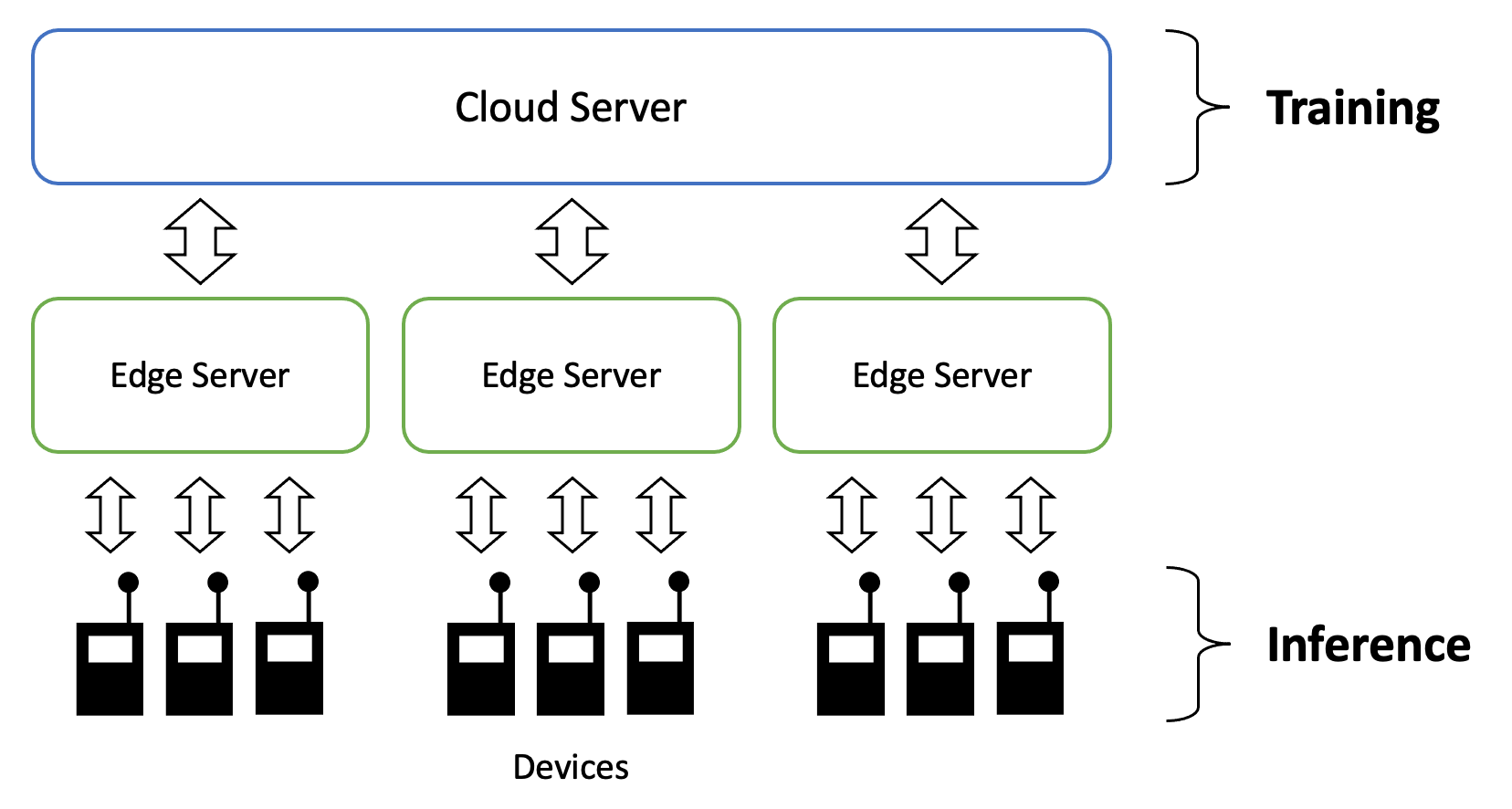}
    \caption{Edge Intelligence (EI). \textcolor{black}{In EI, data gathered from various devices is not processed entirely in the cloud but rather locally on each device. These devices, like sensors in industrial settings, face limitations that make it difficult to deploy large-scale deep learning models typically trained in the cloud, due to constraints such as computing resources and power consumption.}}
    \label{fig:ei}
\end{figure}

However, deep learning models, which have a vast number of parameters, demand considerable computational resources for inference, thereby limiting their practical and sustainable use.
For example, in Edge Intelligence (EI) \cite{cao2020overview,zhou2019edge}, data from individual devices are processed both in the cloud and locally on each device (Fig.\,\ref{fig:ei}).
EI devices, such as sensors in factories, have limited computational resources and power consumption constraints.
This limitation poses a challenge for performing inference using deep learning models with numerous parameters.
Therefore, it is crucial to achieve optimal performance using models with fewer parameters and reduced computational costs in EI applications.

\textcolor{black}{
When considering the application of deep learning models in EI, 
the implementation of model compression techniques is essential \cite{choudhary2020comprehensive, djigal2022machine, murshed2021machine, tekin2023review}.
For example, the techniques include:}
\begin{itemize}
    \item 
\textcolor{black}{Pruning, which reduces the number of parameters in deep learning models \cite{DBLP:journals/corr/HanMD15, lecun1989optimal}.}

    \item 
\textcolor{black}{Quantization, which reduces the number of bits used to represent the weights and activations in deep learning models \cite{gholami2022survey}.}

    \item 
\textcolor{black}{Knowledge distillation, involving training a smaller student model to replicate a larger teacher model  \cite{gou2021knowledge,  hinton2015distilling, hao2023revisit, huang2023knowledge}.}
 
\end{itemize}
\textcolor{black}{
 However, the effectiveness of these
model compression techniques in deep models that incorporate SSMs remains unclear.}

Thus, our goal is to \textcolor{black}{provide a novel and effective model compression method tailored for} S4 models with DSS layers, especially for EI scenario deployment.
To achieve this, we leverage SSM's use in DSS layers to enable the use of various well-established reduction methods \cite{antoulas2005approximation, astolfi2010model, gugercin2008h_2, moore1981principal, sato2018riemannian, sato2019riemannian, sato2023reduced}.
In this study, we employ the balanced truncation method \cite{moore1981principal}, a widely used control theory approach.
To train the original model, we use Skew-HiPPO initialization \cite{gu2022parameterization, gupta2022diagonal}, consistently outperforming models started with random initialization.

The contributions of this paper are summarized as follows:
\textcolor{black}{To reduce computational costs during inference,} we introduce a novel \textcolor{black}{model compression} method that applies the balanced truncation technique to DSS layers in \textcolor{black}{pre-trained S4} models. 
\textcolor{black}{Moreover, we propose using the reduced model parameters obtained by the balanced truncation as initial parameters of S4 models with DSS layers during the main training process.}
As demonstrated in Section \ref{sec:experiments}, 
\textcolor{black}{our trained models combined with the balanced truncation achieved superior accuracy on LRA tasks compared to conventionally trained models using Skew-HiPPO initialization as described in \cite{gu2022parameterization, gupta2022diagonal}, even with fewer parameters.
}
While \cite{massaroli2024laughing} reports minimal impact from dimension reduction in the MultiHyena variant of the Hyena model on performance, our findings with the S4 model underscore a critical distinction with significant performance improvement.

The paper is organized as follows.
In Section \ref{sec:background}, we introduce the balanced truncation method for the reduction of state space models and explain the HiPPO matrix used in Skew-HiPPO initialization.
In Section \ref{sec:model}, we present a deep learning model with DSS layers.
Section \ref{sec:limitation} describes existing training methods for this model, and discusses the model's computational cost during inference.
To address the issue of computational cost, in Section \ref{sec:proposed}, we propose a \textcolor{black}{model compression} method using the balanced truncation method for SSMs.
In Section \ref{sec:experiments}, the results of numerical experiments are presented.
Finally, in Section \ref{sec:conclusion}, we discuss the effectiveness of the proposed method based on the results of the numerical experiments\textcolor{black}{, clarify the limitations of our work,} and outline future work.

\textbf{Notation}
\begin{itemize}
    \item $\mathbb{R}$ and $\mathbb{C}$ denote the sets of real and complex numbers, respectively.

    \item For $a\in \mathbb{C}$, $|a|$ denotes the absolute value of $a$.

    \item For $v=(v_i)\in \mathbb{C}^N$, $\|v\|$ denotes the Euclidean norm of $v$, i.e., $\|v\|:=\sqrt{|v_1|^2+\cdots +|v_N|^2}$.
    
    \item $A^{\top}$ and $A^{*}$ denote the transpose and complex conjugate transpose of matrix $A \in \mathbb{C}^{n\times n}$, respectively.
    \item $f(x)\mid_{x\leq a}$ denotes the function that coincides with $f(x)$, a function defined on $x\geq 0$, on its domain $[0, a]$.
    \item $\mathrm{deg}(f)$ represents the degree of polynomial $f(x)$.
    \item $\mathrm{i}$ denotes the imaginary unit.
    \item $\mathrm{diag}(\lambda_1,\cdots,\lambda_N)\in \mathbb{C}^{N\times N}$ is a diagonal matrix with $\lambda_1,\cdots,\lambda_N\in \mathbb{C}$ as diagonal elements.
    \item $\mathrm{Re}(\alpha)$ and $\mathrm{Im}(\alpha)$ are the real and imaginary parts of $\alpha\in \mathbb{C}$, respectively.
    \item $a*b$ represents the Hadamard product of vectors $a$ and $b$.
    \item $\mathbb{I}_{[a,b]}(x)=\begin{cases}
        1 & \text{if}\quad a\leq x \leq b\\
        0 & \text{otherwise}
    \end{cases}$
    \item $\mathcal{U}(a,b)$ denotes the uniform distribution on $(a, b)$.
    \item $\mathcal{N}(\mu,\sigma)$ is a normal distribution with mean $\mu$ and variance $\sigma$.
\end{itemize}

\section{Preliminaries} \label{sec:background}

In this section, we introduce a state space model (SSM), an important component of the DSS layer. We also present the balanced truncation method \cite{moore1981principal}, a reduction method for SSMs employed in our training method.
Furthermore, we explain the HiPPO matrix \cite{gu2022efficiently} utilized in Skew-HiPPO initialization \cite{gu2022parameterization,
gupta2022diagonal}, which enhances the performance of trained models.

\subsection{State space model (SSM)} \label{subsec:ssm}
A crucial component of a deep learning model discussed in our study, as outlined in Section \ref{sec:model}, is the hidden layer known as the DSS layer. This layer is defined by using the SSM
\begin{align} \label{eq:c-SSM}
   \begin{cases}
    \frac{\mathrm{d}x}{\mathrm{d}t}(t)=Ax(t) +Bu(t) \\
    y(t)=Cx(t),  
   \end{cases}
\end{align}
where $u(t)\in \mathbb{R}$, $y(t)\in \mathbb{C}$, and $x(t)\in \mathbb{C}^N$ denote the input, output, and state, respectively, and $(A, B, C)\in \mathbb{C}^{N\times N} \times \mathbb{C}^{N \times 1} \times \mathbb{C}^{1 \times N}$.
\textcolor{black}{The matrix $A$ denotes a state transition matrix, which describes the internal influence on the time evolution of the internal state $x$.
The matrix $B$ is an input matrix, which describes how the external input $u$ affects the internal state $x$.
The matrix $C$ serves as an output matrix, which describes how the internal state $x$ is transformed into the observable output $y$.
}

Notably, SSM \eqref{eq:c-SSM} is a single-input and single-output system, and the matrices $A$, $B$, $C$, state $x(t)$, and output $y(t)$ are all complex-valued.
Moreover, the state dimension $N$ is relatively large, unlike the standard setting in control theory \cite{dullerud2000course}.

\subsection{Balanced truncation method} \label{subsec:balred}

To address computational issues arising from the large state dimension of SSM \eqref{eq:c-SSM}, we consider using the balanced truncation method \cite{moore1981principal}, a reduction technique.
This method focuses on the controllability and observability of SSM \eqref{eq:c-SSM} to derive another SSM with dimension $r~(\leq N)$, which gives almost the same output as \eqref{eq:c-SSM}:
\begin{align}
\begin{cases}
    \frac{\mathrm{d}x_r}{\mathrm{d}t}(t)=A_rx_r(t)+B_ru(t)\\
    y_r(t)=C_rx_r(t), 
\end{cases}\label{eq:r-dss}
\end{align}
where $A_r\in\mathbb{C}^{r\times r},B_r\in\mathbb{C}^{r \times 1},C_r\in\mathbb{C}^{1 \times r}$.
Below is a brief explanation of the balanced truncation method.
\textcolor{black}{For more detailed information, refer to Appendix \ref{appedix_BT}.}

For SSM (\ref{eq:c-SSM}) with asymptotic stability, where all the real parts of the eigenvalues of matrix $A$ are negative, the controllability Gramian $P$ and observability Gramian $Q$ are defined as
\begin{align}
    P \coloneqq \int_{0}^{\infty} \exp(At)BB^{*}\exp(A^{*}t)\mathrm{d}t,\\
    Q \coloneqq \int_{0}^{\infty} \exp(A^{*}t)C^{*}C\exp(At)\mathrm{d}t.
\end{align}
These are the unique solutions to the Lyapunov equations
\begin{align}
    AP+PA^{*}+BB^{*}=0, \label{Lyap_P}\\
    A^{*}Q+QA+C^{*}C=0, \label{Lyap_Q}
\end{align}
as shown in \cite[Theorem 4.1]{dullerud2000course}.

In the balanced truncation method, the subspace spanned by eigenvectors corresponding to small eigenvalues of the controllability Gramian $P$ and the observability Gramian $Q$ is ignored.
The minimum input energy required to achieve $\lim_{t\to\infty}x(t)=x_f$ from the initial condition $x(0)=0$ is expressed using the controllability Gramian $P$ as
\begin{align}
    \int_{0}^{\infty}\|u(t)\|^2 \mathrm{d}t = x_f^* P^{-1} x_f.
\end{align}
Thus, eigenvectors corresponding to smaller eigenvalues of $P$ correspond to directions in the state space that are less influenced by the input $u$.
On the other hand, when $x(0)=x_0$ and $u(0)=0$, the output energy can be expressed using the observability Gramian $Q$ as
\begin{align}
    \int_{0}^{\infty}\|y(t)\|^2 \mathrm{d}t = x_0^* Q x_0.
\end{align}
Thus, eigenvectors corresponding to smaller eigenvalues of $Q$ correspond to directions in the state space that have less impact on the output $y$.

A coordinate transformation $\bar{x}(t) = Tx(t)$ is applied to SSM \eqref{eq:c-SSM}, obtaining another SSM where the controllability Gramian and observability Gramian coincide as a diagonal matrix $\Sigma$:
\begin{align}
\begin{cases}
    \frac{\mathrm{d}\bar{x}}{\mathrm{d}t}(t)=TAT^{-1}\bar{x}(t)+TBu(t) \\ y(t)=CT^{-1}\bar{x}(t).
\end{cases}
     \label{SSM_change}
\end{align}
SSM \eqref{SSM_change} is referred to as the balanced realization of SSM \eqref{eq:c-SSM}.
Here, the diagonal elements of $\Sigma$ are denoted as $\sigma_1, \cdots, \sigma_N$, which are called the Hankel singular values, satisfying $\sigma_1 \geq \cdots \geq \sigma_N > 0$ under the assumption that SSM \eqref{eq:c-SSM} is controllable and observable.
By partitioning the matrices $TAT^{-1}$, $TB$, $CT^{-1}$ in SSM \eqref{SSM_change} as
\begin{align}
    TAT^{-1} &= \begin{pmatrix}
    A_{11} & A_{12} \\
    A_{21} & A_{22} \\
    \end{pmatrix}, \label{A_trans}\\
    TB &= \begin{pmatrix}
    B_{1} \\
    B_{2} \\
    \end{pmatrix}, \label{B_trans}\\
    CT^{-1} &= \begin{pmatrix}
    C_{1} & C_{2} \\
    \end{pmatrix}, \label{C_trans}
\end{align}
we define
the parameters $(A_r, B_r, C_r)$ of reduced model \eqref{eq:r-dss} as
\begin{align}
(A_r, B_r, C_r) = (A_{11}, B_1, C_1). \label{matrix_reduced}
\end{align}
The resulting system \eqref{eq:r-dss} with \eqref{matrix_reduced} can be interpreted as a reduced SSM of SSM \eqref{eq:c-SSM}, obtained by truncating the state space associated with the smaller Hankel singular values $\sigma_{r+1}, \cdots, \sigma_N$, which correspond to the subspace spanned by eigenvectors that are less influenced by the input or have less influence on the output.
Moreover, if SSM \eqref{eq:c-SSM} is asymptotically stable, the reduced SSM \eqref{eq:r-dss} with \eqref{matrix_reduced} is also asymptotically stable, as shown in \cite[Proposition 4.15]{dullerud2000course}.

\subsection{HiPPO matrix} \label{subsec:hippo}

For the model explained in Section \ref{sec:model}, the parameters of the matrices $(A, B, C)$ in the SSM \eqref{eq:c-SSM} are trained using a suitable optimization algorithm.
The initialization of matrix $A$ significantly influences the performance of trained models, as it sets the initial state for the optimization process.
The High-order Polynomial Projection Operators (HiPPO) matrix \cite{gu2020hippo} is derived from a method for online compression of continuous signals using projections onto subspaces spanned by polynomial bases.
It is well-established that the HiPPO matrix is an effective choice for the initial $A$ \cite{gu2022efficiently, gu2022parameterization, gupta2022diagonal}.

The derivation of the HiPPO matrix is explained below.
With respect to a measure $\mu$ on $[0, \infty)$, let 
\begin{align}
    \begin{split}
    L_2(\mu) \coloneqq \{& f:[0,\infty) \rightarrow {\mathbb C} \Bigm\vert f~\mathrm{is~measurable},\\
    &\int_{0}^{\infty} |f(\tau)|^2 \mathrm{d}\mu(\tau) < \infty \}.
    \end{split}
\end{align}
The inner product and norm on $L_2(\mu)$ are defined as
\begin{align}
    &\langle f_1, f_2 \rangle_\mu \coloneqq \int_{0}^{\infty} f_1^*(\tau)f_2(\tau)\mathrm{d}\mu(\tau),\\
    &\lVert f \rVert _{L_2(\mu)} \coloneqq \langle f, f \rangle_\mu^{1/2},
\end{align}
respectively.

For an input signal $u(t)$ defined on $t \geq 0$, the history $u_{\leq t}\coloneqq u(\tau)\mid_{\tau\leq t}$ at each time $t>0$ is approximated by projecting it onto a subspace spanned by polynomial bases, and the corresponding coefficient vector $x(t)$ represents the history of input signal.
This compression is useful because storing $u_{\leq t}$ requires a significant amount of memory.
Thus, at each time $t$, $x(t)$ contains sufficient information to reconstruct $u_{\leq t}$, even though it requires less memory compared to directly storing $u_{\leq t}$.

The vector $x(t)$ is expressed as the optimal solution to a convex optimization problem,
which is defined by a measure $\mu^{(t)}$ on $(-\infty, t]$ and orthogonal polynomial basis of the subspace of $L_2(\mu^{(t)})$ denoted as $\{g_n^{(t)}\}_{1\leq n \leq N}$ (i.e. $\mathrm{deg}(g_i^{(t)})=i~(i=1,\cdots,N),\langle g_i^{(t)}, g_j^{(t)} \rangle_{\mu^{(t)}} = 0~(i \neq j)$).
The optimization problem is
\begin{align}
    \begin{split}
        &\min_{x(t)\in \mathbb{R}^N } \quad \lVert u_{\leq t}- g \rVert _{L_2(\mu^{(t)})}^2 \\
        &\textrm{subject\,\,to}\quad ~g(\tau) = \sum_{k=1}^{N} x_k(t) g^{(t)}_k(\tau).
    \end{split}
\end{align}
If $\{g_n^{(t)}\}_{1\leq n \leq N}$ is a normalized orthogonal basis, i.e. $\lVert g_i^{(t)} \rVert _{L_2(\mu^{(t)})}=1~(i=1,\cdots,N)$, the optimal solution is given by
\begin{align} \label{eq:x}
    x_k(t) = \langle u_{\leq t}, g^{(t)}_k \rangle_{\mu^{(t)}}.
\end{align}

The vector $x(t)$ defines the approximation $g = \sum_{k=1}^{N} x_k(t) g^{(t)}_k$ of $u_{\leq t}$, thus it retains information necessary for reconstructing the history $u_{\leq t}$ of the input $u(t)$ at time $t$.
This property of memorizing the input history in the state vector will be useful for modeling long sequential data, as capturing dependencies in sequential data requires referencing information from previous inputs to compute each output at every time step.
Moreover, by Equation \eqref{eq:x}, the measure $\mu^{(t)}$ represents the importance of each time step when compressing the history $u_{\leq t}$.
This $x(t)$ satisfies the differential equation
\begin{align} \label{eq:t-ssm}
    \frac{\mathrm{d}x}{\mathrm{d}t}(t)=A(t)x(t)+B(t)u(t),
\end{align}
where $A(t)\in\mathbb{R}^{N\times N},B(t)\in\mathbb{R}^{N \times 1}$ depend on the polynomial basis $\{g^{(t)}_n\}_{1\leq n \leq N}$ and the measure $\mu^{(t)}$.
Unlike SSM \eqref{eq:c-SSM} introduced in Subsection \ref{subsec:ssm}, $A(t)$ and $B(t)$ are time-dependent.

For HiPPO-LegS that is a variant of HiPPO \cite{gu2020hippo}, the measure $\mu^{(t)}$ is defined as the scaled Legendre (LegS) measure
\begin{align}
    \mu^{(t)}= \frac{1}{t}\mathbb{I}_{[0,t]}.
\end{align}
This assigns uniform importance to the entire history $[0, t]$ at each time $t$.
Furthermore, the polynomial basis $g^{(t)}_n(\tau)$ is the normalized orthogonal basis
\begin{align}
    g^{(t)}_n(\tau) = g_n(t,\tau) = (2n+1)^{1/2}P_n\left(\frac{2\tau}{t}-1\right),
\end{align}
where $P_n(\alpha)$ is the Legendre polynomial
\begin{align}
    P_n(\alpha)= \frac{1}{2^n \cdot n!}\frac{\mathrm{d}^n}{\mathrm{d}\alpha^n}(\alpha^2-1)^n.
\end{align}
In this case, as shown in \cite[Appendix D.3]{gu2020hippo}, $x(t)$ satisfies
\begin{align} 
    \label{eq:hippo-dif}
    &\frac{\mathrm{d}x}{\mathrm{d}t}(t)=\frac{1}{t}Ax(t)+\frac{1}{t}Bu(t)\\ 
    \label{eq:hippo-matrix}
    &A_{nk} = -\begin{cases}
    (2n+1)^{1/2}(2k+1)^{1/2} & \text{if}\quad n>k \\
    n+1 & \text{if}\quad n=k \\
    0 & \text{if}\quad n<k
    \end{cases}\\ 
    &B_n = (2n+1)^{1/2}.
\end{align}
This matrix $A$ of Equation \eqref{eq:hippo-matrix} is called the HiPPO matrix.
For the SSM incorporating the HiPPO matrix, the state vector $x(t)$ retains information about the history $u_{\leq t}$ of the input $u$ at each time $t$ \cite{gu2020hippo}.

\section{Deep learning model employed in this study} \label{sec:model}

\begin{figure}[t]
  \centering
  \includegraphics[width=8.5cm]{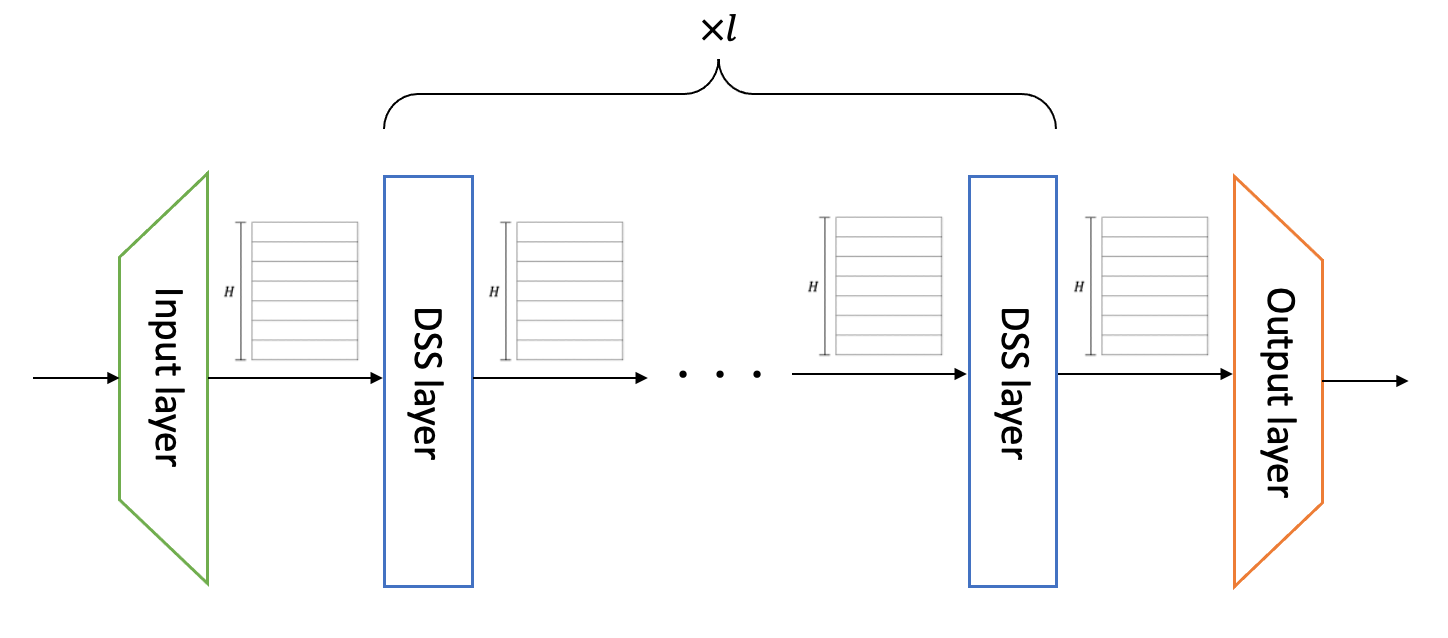}
  \caption{Deep learning model with DSS layers. \textcolor{black}{This represents the overall architecture of the deep learning model used in this study, with the intermediate DSS layer being the most critical component.}}
  \label{fig:whole-model}
\end{figure}

The deep learning model employed in this study, proposed in \cite{gupta2022diagonal}, has the structure illustrated in Fig.\,\ref{fig:whole-model}.
 In this paper, the model is referred to as S4 with DSS layers, despite being named DSS by the authors of \cite{gupta2022diagonal}.
The input layer receives sequential data and outputs $H$ features as 1-dimensional sequential data.
This conversion adapts data from various formats to the DSS layer's input format, as detailed in Subsection \ref{subsec:dss-layer}.
The term $H$ denotes the hidden size, representing the count of features processed by the DSS layer.
Finally, the output layer converts the $H$ features of 1-dimensional sequential data into the model's final output format.
For further details, refer to Appendix \ref{appx:whole-model}.

\subsection{Diagonal State Space layer} \label{subsec:dss-layer}

\begin{figure}[t]
  \centering
  \includegraphics[width=8.5cm]{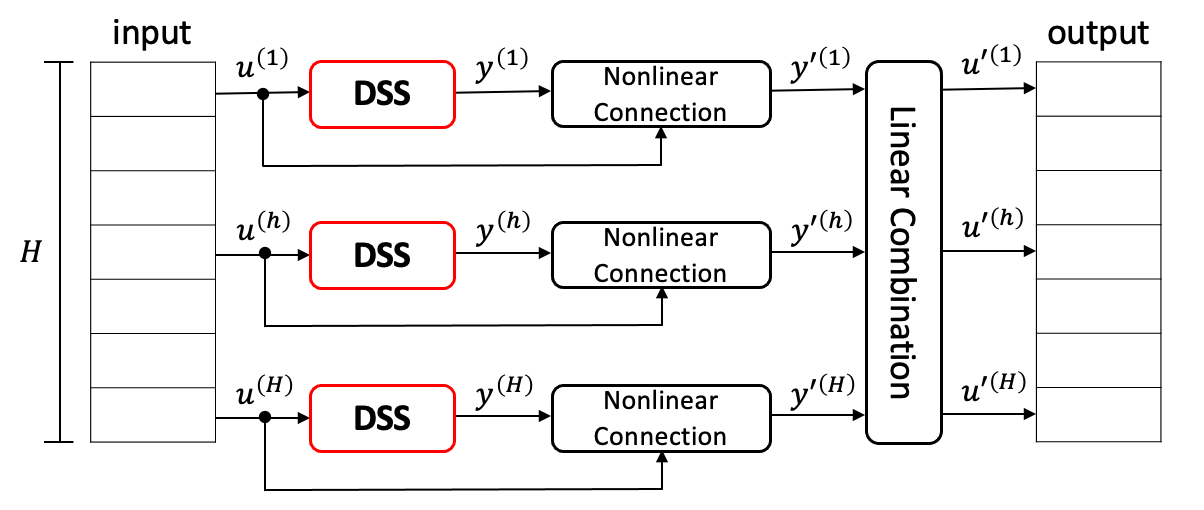}
  \caption{DSS layer\textcolor{black}{, which consists of $H$ DSS models, nonlinear connection blocks, and a linear combination block}.}
  \label{fig:dss-layer}
\end{figure}

The most important component of the deep learning model illustrated in Fig.~\ref{fig:whole-model} is the DSS layer.
As shown in Fig.~\ref{fig:dss-layer}, the DSS layer consists of:
\begin{itemize}
    \item Independent $H$ DSS models.
    \item Nonlinear connection blocks.
    \item Linear combination block.
\end{itemize}
The details of each of these components are explained below.

By restricting the matrix $A \in \mathbb{C}^{N \times N}$ in \eqref{eq:c-SSM} to be diagonal, assuming that the diagonal elements do not lie on the imaginary axis, 
the DSS model is defined by the following discretization for a sample time $\Delta \in \mathbb{R}_{>0}$:
\begin{align}\label{eq:d-SSM}
    \begin{cases}
    x_{k+1} = \bar{A} x_{k} + \bar{B} u_k \\
    y_k = \bar{C} x_k,
    \end{cases}
\end{align}
where $\bar{A} = e^{A\Delta}$, $\bar{B} = (\bar{A}-I)A^{-1}B$, and $\bar{C} = C$. 
The diagonal elements of the matrix $\bar{A}$ do not lie on the unit circle in the complex plane, due to the assumption on the matrix $A$.

The nonlinear connection block receives the input $u_k$ and output $y_k$ from the DSS model and outputs 1-dimensional sequential data
\begin{align} \label{eq:residual}
    y'_k = \mathrm{GELU}(\mathrm{Re}(y_k) + Du_k),
\end{align}
where $D \in \mathbb{R}$, and GELU\cite{hendrycks2016gaussian} is a nonlinear activation function expressed as
\begin{align}
    \mathrm{GELU}(\alpha)=\alpha\Phi(\alpha).
\end{align}
Here, $\Phi(\alpha)$ is the cumulative distribution function of the standard normal distribution.
This approach is expected to enhance the performance of the model.
Note that $\mathrm{Re}(y_k)$ is used in Equation \eqref{eq:residual} since $y_k$ can be a complex number.

Finally, the $H$ 1-dimensional sequential data $(y_k'^{(h)})_{1\leq h \leq H}$ outputted from each nonlinear connection block is mixed to obtain the final output of the DSS layer, resulting in $H$ 1-dimensional sequential data $(u_k'^{(h)})_{1\leq h \leq H}$.
With parameters of weight $W_{out} \in \mathbb{R}^{H \times H}$ and bias $b \in \mathbb{R}^H$, the output is expressed as
\begin{align} \label{linear_com}
    u_k'^{(h)} = \sum_{h'=1}^{H} W_{out,hh'} y_k'^{(h')} + b_h\cdot\bm{1},
\end{align}
where $\bm{1}$ is a 1-dimensional sequential data of the same length as $y_k'^{(h)}$ with all elements $1$.

\subsection{DSS${}_\text{EXP}$ and DSS${}_\text{SOFTMAX}$} \label{subsec:dss-variants}

The output of DSS \eqref{eq:d-SSM} can be calculated as
\begin{align}\label{eq:convolution}
    y_k = \sum_{i=0}^{k} h_i\cdot u_{k-i}
\end{align}
with $h_i\coloneqq \bar{C}\bar{A}^i\bar{B}$, which is referred to as the impulse response of DSS \eqref{eq:d-SSM}.
Given a sample time $\Delta$, $h_i$ is determined by $(A,B,C)$, and different sets of parameters $(A,B,C)$ may result in the same sequence of impulse responses.
In fact, 
\begin{align} \label{eq:SSM-kernel}
    \begin{split}
    \bar{K}_{\Delta,L}(A,B,C) 
    &\coloneqq (h_0,h_1,\cdots,h_{L-1})\\
    \end{split}
\end{align}
are the same for different $(A,B,C)$, as described below \cite{gupta2022diagonal}.

\begin{prop} \label{prop:SSM-kernel}
Suppose that the parameters $A=\mathrm{diag}(\lambda_1,\cdots,\lambda_N),B,C,\Delta$ of DSS \eqref{eq:d-SSM} are given, and define $K := \bar{K}_{\Delta,L}(A,B,C) \in \mathbb{C}^{L}$.
Then, there exist $w,\Tilde{w}\in \mathbb{C}^{1 \times N}$ satisfying the following equations:
\begin{enumerate}[(a)]
    \item $K =\bar{K}_{\Delta,L}(A,(1)_{1\leq i \leq N},\Tilde{w})$\\
        $\quad = \Tilde{w} \cdot A^{-1}(e^{A\Delta}-I) \cdot \mathrm{elementwise\mathchar`-exp}(P)$
    \item $K =\bar{K}_{\Delta,L}(A,((e^{L\lambda_i\Delta}-1)^{-1})_{1\leq i \leq N},w)$\\
    $\quad = w \cdot A^{-1} \cdot \mathrm{row\mathchar`-softmax}(P)$
\end{enumerate}
where
\begin{align}
    &\mathrm{elementwise\mathchar`-exp}(P) = \left(\exp(P_{ik})\right)_{1\leq i \leq N, 0\leq k < L}, \\
    &\mathrm{row\mathchar`-softmax}(P) = \left(\frac{\exp(P_{ik})}{\Sigma_{r=0}^{L-1} \exp(P_{ir})}\right)_{1\leq i \leq N, 0\leq k < L},\\
    &P_{i,k}=\lambda_i k\Delta.
\end{align}
\end{prop}

Proposition \ref{prop:SSM-kernel} implies that, under weak assumptions, the impulse responses $h_0,h_1,\cdots,h_{L-1}$ of DSS \eqref{eq:d-SSM} can be achieved with special structure of $(A,B,C)$.
DSS \eqref{eq:d-SSM} with $(B,C)=((1)_{1\leq i \leq N},w)$, as stated in Proposition \ref{prop:SSM-kernel}(a), is referred to as DSS${}_\text{EXP}$, and DSS \eqref{eq:d-SSM} with $(B,C)=(((e^{L\lambda_i\Delta}-1)^{-1})_{1\leq i \leq N},w)$, as stated in Proposition \ref{prop:SSM-kernel}(b), is referred to as DSS${}_\text{SOFTMAX}$ \cite{gupta2022diagonal}.
DSS${}_\text{EXP}$ and DSS${}_\text{SOFTMAX}$ offer different approaches to modeling the impulse responses, with potential implications for the performance and interpretability of the DSS model.
In the following sections, we utilize DSS${}_\text{EXP}$ or DSS${}_\text{SOFTMAX}$ as DSS \eqref{eq:d-SSM}.

\section{Existing training methods and limitations} \label{sec:limitation}

In the training of deep learning models, the goal is to minimize the loss function $E(W)$ with respect to the training dataset $\{(\chi_i, d_i)\}$.
Here, $(\chi_i, d_i)$ represents a pair of input $\chi_i$ and its desired output $d_i$, and $W$ denotes the parameters of the model.
The model's output for an input $\chi$ with parameters $W$ is denoted as $\zeta(\chi;W)$.
For each input $\chi_i$, a loss function $E_i(W)$ is defined to  measure the difference between the desired output $d_i$ and the model's output $\zeta(\chi_i;W)$.
The loss function for the entire training dataset is expressed as $E(W)=\Sigma_{i}E_i(W)$, where the summation is over all training examples.
As an algorithm for minimizing the loss function $E(W)$, we can consider using AdamW \cite{loshchilov2017decoupled}.

\subsection{Training parameters within the Diagonal State Space layer} \label{subsec:parameter}

Among the parameters $W$ trained in our deep learning model, those in the DSS layer include:
\begin{itemize}
    \item $(A, B, C, \Delta)$ for each of the $H$ DSS models, as defined in  \eqref{eq:d-SSM}.
    \item $D$ for each of the $H$ nonlinear connection blocks, as defined in  \eqref{eq:residual}.
    \item Weight $W_{out} \in \mathbb{R}^{H \times H}$ and bias $b \in \mathbb{R}^H$ for the linear combination block, as defined in \eqref{linear_com}.
\end{itemize}

For DSS${}_\text{EXP}$ defined in Subsection \ref{subsec:dss-variants}, the parameters $(A, B, C)$ are defined as
\begin{align} \label{eq:dss-exp}
    \begin{split}
    &A = \mathrm{diag}(\lambda_1,\cdots,\lambda_N), \\
    &B = (1)_{1\leq i \leq N},\\
    &C = w,
    \end{split}
\end{align}
where
\begin{align} \label{eq:lambda-exp}
    \lambda_i=-\exp(\Lambda_{re,i})+\mathrm{i}\cdot \Lambda_{im,i}.
\end{align}
For DSS${}_\text{EXP}$, the parameters $\Lambda_{re},\Lambda_{im}\in \mathbb{R}^{N}$ and $w\in \mathbb{C}^{N}$ are trained to determine $(A,B,C)$.

For DSS${}_\text{SOFTMAX}$ defined in Subsection \ref{subsec:dss-variants}, the parameters $(A, B, C)$ are defined as:
\begin{align} \label{eq:dss-softmax}
    \begin{split}
    &A = \mathrm{diag}(\lambda_1,\cdots,\lambda_N), \\
    &B = ((e^{L\lambda_i\Delta}-1)^{-1})_{1\leq i \leq N},\\
    &C = w,
    \end{split}
\end{align}
where
\begin{align} \label{eq:lambda-softmax}
    \lambda_i= \Lambda_{re,i}+\mathrm{i}\cdot \Lambda_{im,i}.
\end{align}
Similarly to DSS${}_\text{EXP}$, the parameters $\Lambda_{re},\Lambda_{im}\in \mathbb{R}^{N}$ and $w\in \mathbb{C}^{N}$ are trained to determine $(A,B,C)$ for DSS${}_\text{SOFTMAX}$, with a different expression for $A$ and $B$.

\subsection{Initialization of the DSS Layer} \label{subsec:initialization}

The performance of S4 with DSS layers is sensitive to initialization of the state matrix $A$.
To obtain an effective initial value for $A$, the HiPPO matrix is decomposed into a normal matrix and a low-rank matrix.
This decomposition allows for a more structured and interpretable initialization of the state matrix $A$, which can improve the performance of the model.
The eigenvalues of the normal matrix are employed to initialize the diagonal elements of $A$.

In more detail, the HiPPO matrix $\mathcal{H} \in \mathbb{R}^{N'\times N'}$ is defined as explained in Subsection \ref{subsec:hippo}:
\begin{align}
    \mathcal{H}_{nk} = -\begin{cases}
    (2n+1)^{1/2}(2k+1)^{1/2} & \text{if}\quad n>k \\
    n+1 & \text{if}\quad n=k \\
    0 & \text{if}\quad n<k.
    \end{cases}
\end{align}
For SSM (\ref{eq:c-SSM}) incorporating this HiPPO matrix, the state $x(t)$ retains information about the history of the input $u(t)$ \cite{gu2020hippo}.
The HiPPO matrix $\mathcal{H}$ can be decomposed into a normal matrix and a low-rank matrix as
\begin{align}
    \mathcal{H} = \mathcal{H}' - \frac{1}{2}PQ^{\top},
\end{align}
where $\mathcal{H}'\in \mathbb{R}^{N'\times N'},P\in \mathbb{R}^{N'},Q\in \mathbb{R}^{N'}$ are defined as
\begin{align}
    &\mathcal{H}'_{nk} = -\begin{cases}
    (2n+1)^{1/2}(2k+1)^{1/2}/2 & \text{if}\quad n>k \\
    1/2 & \text{if}\quad n=k \\
    -(2n+1)^{1/2}(2k+1)^{1/2}/2 & \text{if}\quad n<k
    \end{cases}\\
    &P_n = (2n+1)^{1/2},\quad Q_k = (2k+1)^{1/2}.
\end{align}
This $\mathcal{H}'$ is a normal matrix.
Under the assumption that
$N'=2N$ and $\mu_1, \ldots, \mu_N$ are the eigenvalues of $\mathcal{H}'\in \mathbb{R}^{2N\times 2N}$ with positive imaginary parts,
 $\Lambda_{re}, \Lambda_{im} \in \mathbb{R}^{N}$ are defined as follows:
\begin{itemize}
    \item For DSS${}_\text{EXP}$,
\begin{align}
    \Lambda_{re,i} = \log(-\mathrm{Re}(\mu_i)),\quad \Lambda_{im,i} = \mathrm{Im}(\mu_i).
\end{align}    

\item 
For DSS${}_\text{SOFTMAX}$,
\begin{align}
    \Lambda_{re,i} = \mathrm{Re}(\mu_i),\quad \Lambda_{im,i} = \mathrm{Im}(\mu_i).
\end{align}
\end{itemize}

 Using $\Lambda_{re}$ and $\Lambda_{im}$, the matrix $A$ is initialized as $A \coloneqq \mathrm{diag}(\lambda_1, \ldots, \lambda_N)$, where each $\lambda_i$ is derived from Equation \eqref{eq:lambda-exp} for DSS${}_\text{EXP}$ or Equation \eqref{eq:lambda-softmax} for DSS${}_\text{SOFTMAX}$. This process is known as the Skew-HiPPO initialization \cite{gu2022parameterization, gupta2022diagonal}. Other parameters within the DSS are randomly sampled, as detailed in Section \ref{sec:experiments}. According to \cite{gupta2022diagonal}, models utilizing Skew-HiPPO initialization demonstrate superior prediction accuracy compared to those initialized with values from $\mathcal{N}(0, 1)$.

\subsection{Computational cost of the DSS layer output} \label{subsec:whole-compute}

When the input is entered one by one or all at once, reducing $H$ (the hidden size) and $N$ (the state dimension) facilitates a reduction in the computational cost of the DSS layer output.
In fact, the time and space complexities of the DSS layer output are as illustrated in Table~\ref{table:computation-step} and Table~\ref{table:computation-whole}, respectively, as discussed below.
Here, the input length of the 1-dimensional sequence is denoted as $L$.

In the case where input $u_k$ at each time step is entered one by one into DSS \eqref{eq:d-SSM}, the output $y_k$ can be computed using the previous state vector $x_{k-1}$ according to \eqref{eq:d-SSM}.
The time and space complexities per step are both $O(N)$.
The time complexity for processing the entire input of length $L$ is $O(NL)$, and the space complexity involves overwriting at each step, thus remaining $O(N)$.
For the nonlinear connection block, both the time and space complexities per step are $O(1)$.
The time complexity for processing the entire input of length $L$ is $O(L)$, and the space complexity involves overwriting at each step, thus remaining $O(1)$.
Regarding the following linear combination block, the time complexity per step is $O(H^2)$, and the space complexity is $O(H)$.
The time complexity for processing the entire input of length $L$ is $O(H^2L)$, and the space complexity involves overwriting at each step, thus remaining $O(H)$.
Therefore, adding up $H$ DSS, $H$ nonlinear connection blocks, and one linear combination block, the time complexity of the DSS layer output when input is entered one by one is $O(HNL) + O(H^2L)$, and the space complexity is $O(HN)$ (Table~\ref{table:computation-step}).

In the case where whole input $(u_k)_{0\leq k < L}$ is entered all at once, the output
\begin{align}\label{eq:kernel-convolution}
    y_k = \sum_{i=0}^{L-1} h_i\cdot u_{k-i}
\end{align}
can be efficiently computed.
In fact,
leveraging the fast Fourier transform \cite{cormen2009introduction} implies that the time complexity is $O(L\log(L))$ and the space complexity is $O(L)$.
Furthermore, the computation is parallelizable.
Besides, the impulse response $h_i$ in Equation (\ref{eq:kernel-convolution}) can be easily computed.
In fact, for a diagonal matrix $A\in\mathbb{C}^{N\times N}$ with diagonal elements $\lambda_i$, $h_i$ can be calculated as
\begin{align}
    \begin{split}
    h_i &= \bar{C}\bar{A}^i\bar{B} \\
    &= Ce^{A\cdot i\Delta}(e^{A\Delta}-I)A^{-1}B\\
    &= \sum_{j=1}^{N}C_j \cdot e^{\lambda_j\cdot i\Delta}(e^{\lambda_j\Delta}-1)\lambda_j^{-1}\cdot B_j.  
    \end{split}
\end{align}
The computation time for the sequence of impulse responses $(h_0,h_1,\cdots,h_{L-1})$ is $O(NL)$.
As for the nonlinear connection block, the time and space complexities are both $O(L)$
Regarding the following linear combination block, the time complexity is $O(H^2L)$, and the space complexity is $O(HL)$.
Therefore, adding up $H$ DSS, $H$ nonlinear connection blocks, and one linear combination block, the time complexity of the DSS layer output when input is entered all at once is $O(HL\log(L))+O(H^2L)$, and the space complexity is $O(HL)$ (Table~\ref{table:computation-whole}).

Additionally, Equation \eqref{eq:kernel-convolution} is an approximation that holds when DSS (\ref{eq:d-SSM}) is asymptotically stable and $L$ is sufficiently large.
The exact output of DSS \eqref{eq:d-SSM} is given by \eqref{eq:convolution}. However, when \eqref{eq:d-SSM} is asymptotically stable and $L$ is sufficiently large, $h_i \approx 0$ for $i \geq L$, making the approximation in \eqref{eq:kernel-convolution} valid.

\begin{table}[t]
    \caption{Computational complexities of the DSS layer output when input of length $L$ is entered one by one}
    \label{table:computation-step}
    \centering
     \begin{tabular}{ccc}
       \hline
       & Time & Space \\
       \hline \hline
       DSS (1 unit) & $O(NL)$ & $O(N)$ \\
       Nonlinear connection(1 unit) & $O(L)$ & $O(1)$  \\
       Linear combination & $O(H^2L)$ & $O(H)$  \\
       \hline
       Total & $O(HNL)+O(H^2L)$ & $O(HN)$  \\
       \hline
     \end{tabular}
   \end{table}

   \begin{table}[t]
    \caption{Computational complexities of the DSS layer output when input of length $L$ is entered at once}
    \label{table:computation-whole}
    \centering
     \begin{tabular}{ccc}
       \hline
       & Time & Space \\
       \hline \hline
       DSS (1 unit) &  $O(L\log(L))$ &  $O(L)$ \\
       Nonlinear connection(1 unit)  & $O(L)$ & $O(L)$  \\
       Linear combination & $O(H^2L)$ & $O(HL)$  \\
       \hline
       Total & $O(HL\log(L))+O(H^2L)$ & $O(HL)$  \\
       \hline
     \end{tabular}
   \end{table}

\subsection{Issues for practical applications} \label{subsec:issue}

Let us consider the issues that hinder the application of S4 with DSS layers to EI \cite{cao2020overview, zhou2019edge}, as explained in Section \ref{sec:intro}.
These issues include memory constraints, computational complexity, and the trade-offs between model performance and resource efficiency.

The following can be concluded from the arguments of Subsection \ref{subsec:whole-compute}:
\begin{itemize}
    \item When processing the entire input of large length $L$ at once, the application of S4 with DSS layers to EI is challenging, even if $H$ and $N$ of the trained model are sufficiently small.
    This is because the space complexity
is $O(HL)$ (Table\,\ref{table:computation-whole}), making it difficult to conduct inference in devices with small capacity memory (e.g. sensors set in factories).

\item  When processing the input one-by-one, which corresponds to $L=1$, S4 with DSS layers can be applied to EI if $H$ and $N$ of the trained model are sufficiently small. 
Specifically, the time and space complexities are those shown in Table ~\ref{table:computation-step}.
That is, when $L=1$, the time complexity is $O(HN)+O(H^2)$, and the space complexity is $O(HN)$.
This implies that even for very large input sequences, inference can be conducted in devices with small capacity memory.
\end{itemize}

In summary, to apply S4 with DSS layers to EI, the input needs to be processed one-by-one, and it is desirable to keep the values of $H$ and $N$ as small as possible.
However, excessively small values of $H$ and $N$ may limit the model's capacity to capture complex patterns in the data, leading to a deterioration in performance.

\section{Proposed Model Compression Method} \label{sec:proposed}

In this section,
to address the issues discussed in Subsection \ref{subsec:issue},
we propose an effective \textcolor{black}{model compression} method for S4 with DSS layers aiming to reduce computational costs during inference by one-by-one processing.
Specifically, this method enables the acquisition of parameter values that achieve higher accuracy compared to existing methods when training models with DSS layers of the same $H$ and $N$.

The following procedure is our 
 proposed \textcolor{black}{model compression} method.
 \begin{enumerate}
     \item  Apply the balanced truncation method, as explained in Subsection \ref{subsec:balred}, to a large-scale DSS that is part of a trained model.

     \item  Retrain the model using the reduced DSS obtained in step 1) for improved initialization.
 \end{enumerate}

\begin{figure}[t]
    \centering
    \includegraphics[width=8cm]{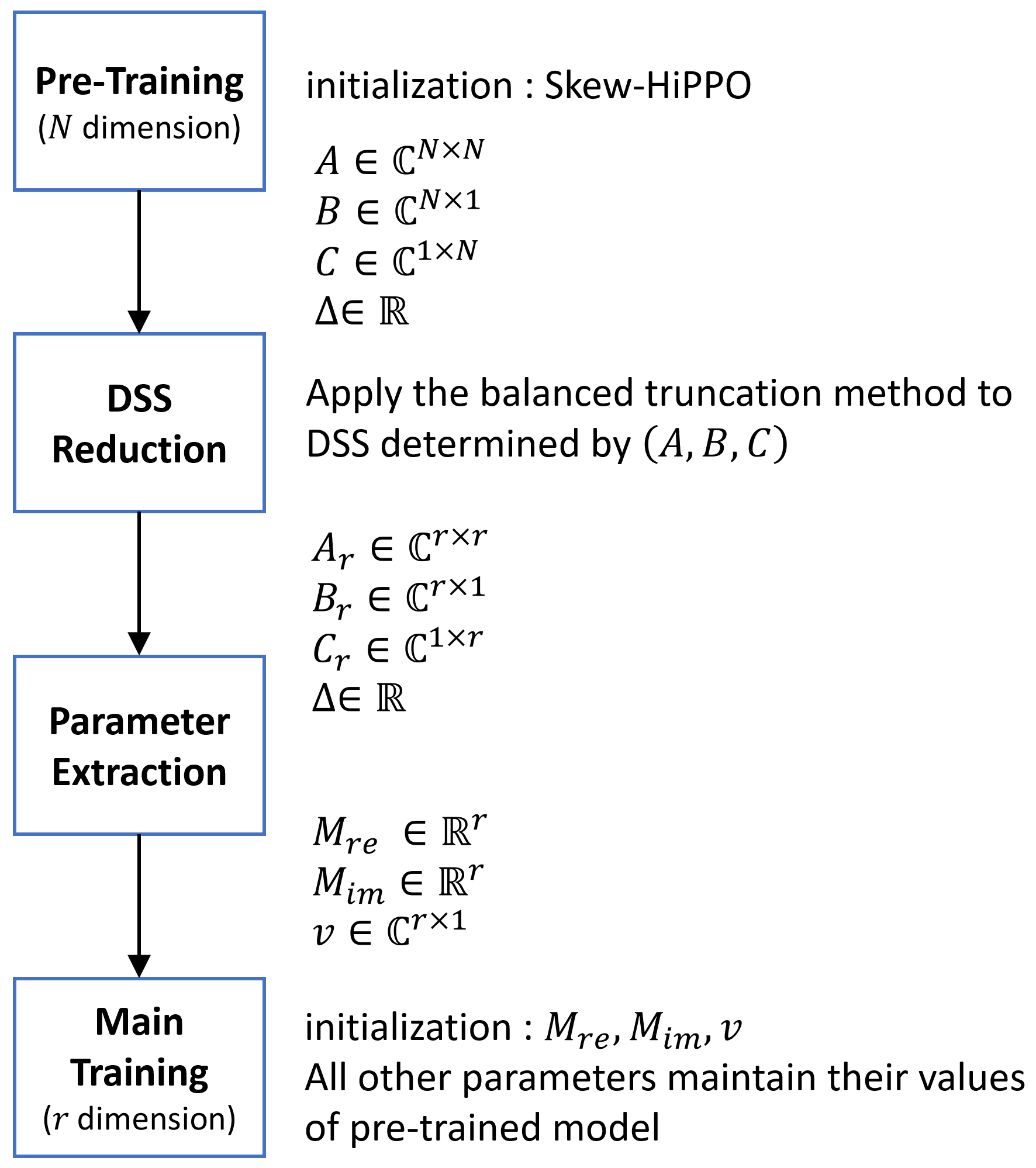}
    \caption{Proposed method\textcolor{black}{, which consists of Pre-Training, DSS Reduction, Parameter Extraction, and Main Training. At DSS Reduction step, we use the balanced truncation method.}}
    \label{fig:flow}
\end{figure}

In more detail, our proposed method consists of the following Pre-Training, DSS Reduction, Parameter Extraction, and Main Training,
illustrated in Fig.\,\ref{fig:flow}.

\subsubsection*{Pre-Training}
The parameters $(A, B, C)$ and $\Delta$ of DSS \eqref{eq:d-SSM} are determined by training the model with the Skew-HiPPO initialization  \cite{gupta2022diagonal}, as detailed in Section \ref{subsec:initialization}.
For DSS${}_\text{EXP}$ and DSS${}_\text{SOFTMAX}$, calculations use \eqref{eq:dss-exp} and \eqref{eq:dss-softmax}, respectively.

\subsubsection*{DSS Reduction}
The DSS model \eqref{eq:d-SSM}, determined by parameters $(A, B, C)$ and $\Delta$ obtained through Pre-Training, is reduced using the balanced truncation method described in Subsection \ref{subsec:balred} \textcolor{black}{and Appendix \ref{appedix_BT}}.
The state dimension is reduced to $r (\leq N)$, resulting in the reduced SSM (\ref{eq:r-dss}).
Here, the sample time $\Delta$ remains unchanged.

\subsubsection*{Parameter Extraction}
Assuming $A_r$ is diagonalizable, we can transform the reduced SSM \eqref{eq:r-dss} into the DSS \eqref{eq:d-SSM}, as detailed below.
There exist a diagonal matrix $M=\mathrm{diag}(\mu_1,\cdots,\mu_{r})$ and an invertible matrix $V\in\mathbb{C}^{r \times r}$ satisfying $A_r = VMV^{-1}$.
Using a coordinate transformation $\hat{x}=V^{-1}x_r$, we obtain the new SSM
\begin{align} \label{reduced_diag}
\begin{cases}
    \frac{\mathrm{d}\hat{x}}{\mathrm{d}t}(t) = M\hat{x}(t) + V^{-1}B_ru(t)\\
    y_r(t)=C_rV\hat{x}(t),
\end{cases}
\end{align}
which
is equivalent to the reduced SSM \eqref{eq:r-dss}.
Consequently, the transformation allows expressing the reduced SSM \eqref{eq:r-dss} in the explicitly diagonal form of DSS, as shown in \eqref{reduced_diag}.

From Proposition \ref{prop:SSM-kernel}, the impulse response of DSS \eqref{reduced_diag} with the state dimension $r$ can also be derived from DSS${}_\text{EXP}$ or DSS${}_\text{SOFTMAX}$.
\begin{itemize}
    \item For DSS${}_\text{EXP}$, the parameters are determined as
\begin{align}
        & M_{re,i} = \log(-\mathrm{Re}(\mu_i)), \\
        & M_{im,i} = \mathrm{Im}(\mu_i), \\
       & v = (C_rV)^{\top} * (V^{-1}B_r).        \label{v1}
\end{align}

\item For DSS${}_\text{SOFTMAX}$, the parameters are determined as
\begin{align}
    &M_{re,i} = \mathrm{Re}(\mu_i), \\
    &M_{im,i} = \mathrm{Im}(\mu_i), \\
    &v = (C_rV)^{\top} * (V^{-1}B_r) * (e^{L\mu_i \Delta}-1)_{1\leq i \leq r}. \label{v2}
\end{align}
\end{itemize}
 For the vectors of \eqref{v1} and \eqref{v2}, refer to the proof of Proposition in \cite[Appendix A.1]{gupta2022diagonal}.

\subsubsection*{Main Training}
As detailed in Subsection \ref{subsec:parameter}, $\Lambda_{re}, \Lambda_{im} \in \mathbb{R}^{r}$ and $w \in \mathbb{C}^{r}$ are training parameters within DSS${}_\text{EXP}$ and DSS${}_\text{SOFTMAX}$.
Here, $(\Lambda_{re}, \Lambda_{im})$ and $w$ are initialized with $(M_{re}, M_{im})$ and $v$ obtained by Parameter Extraction, respectively.
 It is important to note that the dimension, previously denoted as $N$, is adjusted to $r$ for the context of this initialization.
All other parameters maintain their values as obtained from the Pre-Training phase, ensuring consistency in the model's initialization process.

\section{Numerical Experiments} \label{sec:experiments}

To evaluate the proposed method, we employed tasks of LRA \cite{tay2021long}, \textcolor{black}{which is available at \url{https://github.com/google-research/long-range-arena}.}
The benchmark includes sequence data ranging from 1,000 to 16,000 in length and evaluates the model's ability to capture long-range dependencies required for learning long sequences.
In the text classification task, we classify movie reviews in the Internet Movie Database (IMDb) review dataset \cite{maas2011learning} as negative or positive.
\textcolor{black}{Tables \ref{table_negative} and \ref{table_positive} summarize the statistics of the text classification dataset, including the counts and lengths of the raw data sequences. These sequences are truncated or padded as  necessary to ensure consistent input lengths.
}

\textcolor{black}{The experiments were conducted on a machine running Windows 10, equipped with 64 GB of memory and an 11th Gen Intel Core i9-11980HK CPU.
The model training and evaluation code was implemented in Python using PyTorch 1.11.0 and TensorFlow 2.12.0, and executed with the NVIDIA RTX A3000 Laptop GPU.}

\begin{table}[t]
  \centering
    \caption{Text classification (negative)}
    \label{table_negative}
    \begin{tabular}{lcc}
      \hline
       & \textbf{Train} & \textbf{Test} \\ 
      \hline
      Number of examples & 12,500 & 12,500 \\
      Max text length & 8,969 & 6,385 \\
      Min text length & 52 & 32 \\
      Avg text length & 1302.97904 & 1285.14968 \\
      \hline
    \end{tabular}
    \end{table}

\begin{table}[t]
    \centering
    \caption{Text classification (positive)}
    \label{table_positive}
    \begin{tabular}{lcc}
      \hline
       & \textbf{Train} & \textbf{Test} \\ 
      \hline
      Number of examples & 12,500 & 12,500 \\
      Max text length & 13,704 & 12,988 \\
      Min text length & 70 & 65 \\
      Avg text length & 1347.16024 & 1302.43512 \\
      \hline
    \end{tabular}
\end{table}

\subsection{Comparison with existing training methods} \label{subsec:result}

Table\,\ref{table:exp} and Table\,\ref{table:softmax} show the accuracy of models obtained through various training methods, using DSS${}_\text{EXP}$ and DSS${}_\text{SOFTMAX}$ respectively,
where $N$ denotes the dimension of the state vector of DSS.
\textcolor{black}{The number of DSS layers is $4$ and the hidden size $H$ is $16$.
}
The columns labeled
``before'' and ``after'' denote the accuracy of the model with the initial parameter values and the accuracy of the model after training from that initial state, respectively.

In the ``HiPPO'' column on the left, the Skew-HiPPO initialization explained in Subsection \ref{subsec:initialization} was used to initialize the state matrix $A$ of each DSS.
In the middle column ``Random'', the initial values of $A$ were randomly sampled.
For DSS${}_\text{EXP}$, the real and imaginary parts of the diagonal elements of matrix $A$ were sampled from $\mathcal{U}(-1, 0)$ and $\mathcal{N}(0, 1)$, respectively.
For DSS${}_\text{SOFTMAX}$, the real and imaginary parts of the diagonal elements were sampled from $\mathcal{N}(0, 1)$.
Other parameters within DSS were randomly sampled for both ``HiPPO'' and ``Random''.
The real and imaginary parts of each element in $w \in \mathbb{C}^{N}$ were sampled from $\mathcal{N}(0, 1)$.

\begin{table}[t]
    \caption{Accuracy of models through various training methods (Text classification, DSS${}_\text{EXP}$, $H=16$)}
    \label{table:exp}
    \centering
    \begin{tabular}{c c c c c}
        \hline
        $N$ &  & \textbf{HiPPO} & \textbf{Random} & \textbf{Proposed Method} \\
        \hline
        \hline
        \multirow{2}{*}{128} & before & 0.5000 & 0.4998 &  \\
                             & after & 0.7997 & 0.7731 &  \\
        \hline
        \multirow{2}{*}{64}  & before & 0.4967 & 0.4982 &  \\
                             & after & 0.8012 & 0.7968 &  \\
        \hline
        \multirow{2}{*}{32}  & before & 0.5008 & 0.4978 &  \\
                             & after & 0.7990 & 0.8100 &  \\
        \hline
        \multirow{2}{*}{16}  & before & 0.4936 & 0.4970 & 0.8310 \\
                             & after & \textbf{0.8310} & 0.8071 & 0.8396 \\
        \hline
        \multirow{2}{*}{8}   & before & 0.4996 & 0.5028 & 0.5011 \\
                             & after & 0.8182 & 0.8115 & 0.8376 \\
        \hline
        \multirow{2}{*}{4}   & before & 0.4960 & 0.4964 & 0.5002 \\
                             & after & 0.8042 & 0.8155 & \textbf{0.8418} \\
        \hline
    \end{tabular}
\end{table}

\begin{table}[t]
    \caption{Accuracy of models through various training methods (Text classification, DSS${}_\text{SOFTMAX}$, $H=16$)}
    \label{table:softmax}
    \centering
    \begin{tabular}{c c c c c}
        \hline
        $N$ &  & \textbf{HiPPO} & \textbf{Random} & \textbf{Proposed Method}\\
        \hline
        \hline
        \multirow{2}{*}{128} & before & 0.4984 & 0.5000 & 0.8216 \\
                             & after & \textbf{0.8216} & 0.7540 & 0.8359 \\
        \hline
        \multirow{2}{*}{64}  & before & 0.4993 & 0.5000 & 0.5013 \\
                             & after & 0.8158 & 0.7544 & 0.8382 \\
        \hline
        \multirow{2}{*}{32}  & before & 0.4994 & 0.4999 & 0.5000 \\
                             & after & 0.8026 & 0.7496 & 0.8370 \\
        \hline
        \multirow{2}{*}{16}  & before & 0.5011 & 0.4994 & 0.5005 \\
                             & after & 0.8190 & 0.7461 & 0.8402 \\
        \hline
        \multirow{2}{*}{8}   & before & 0.5000 & 0.5072 & 0.5036 \\
                             & after & 0.8071 & 0.7722 & \textbf{0.8412} \\
        \hline
        \multirow{2}{*}{4}   & before & 0.4940 & 0.5000 & 0.5012 \\
                             & after & 0.7948 & 0.7819 & 0.8377 \\
        \hline
    \end{tabular}
\end{table}

For DSS${}_\text{SOFTMAX}$, it has been reported in \cite{gupta2022diagonal} that ``HiPPO'' using Skew-HiPPO initialization achieves higher accuracy after training compared to ``Random'' using randomly sampled initial values.
The results in Table~\ref{table:softmax} are cosistent, where ``HiPPO'' achieves higher accuracy after training compared to ``Random'' for each $N$, while each accuracy before training is near.
For DSS${}_\text{EXP}$, the same trend was observed for almost all $N$ as shown in Table~\ref{table:exp}.

The "Proposed Method" column on the right describes our approach, which uses a reduced SSM from the Pre-Trained models with $N=16$ $({\text{DSS}}_\text{EXP})$ and $N=128$ $({\text{DSS}}_\text{SOFTMAX})$, obtained through the balanced truncation method, to initialize Main Training.
\textcolor{black}{In Table \ref{table:exp}, the "Proposed Method" entries for $N=32$, $N=64$, and $N=128$ are blank, because the balanced truncation does not permit expanding the state dimension beyond the original size of $N=16$.}

 Before Main Training,
the accuracy of models  using the ``Proposed Method'' is comparable to those using ``Random'' and ``HiPPO'' for each $N$ excluding $N=16$ for DSS${}_\text{EXP}$ and $N=128$ for DSS${}_\text{SOFTMAX}$.
However, after Main Training, the accuracy of ``Proposed Method'' exceeded that of ``HiPPO'' for each $N$.
This result is noteworthy because the ``Proposed Method'' tends to outperform ``HiPPO'' after Main Training,
despite having similar accuracies before the training.

The following points are particularly noteworthy.
\begin{itemize}
    \item For DSS${}_\text{EXP}$ shown in Table \ref{table:exp}, the highest accuracy after Main Training with ``Proposed Method'' was 0.8418 at $N=4$.
Notably, this exceeded the accuracy after training with ``HiPPO'' at $N=16$, which was 0.8310, despite having a smaller $N$ while maintaining the same hidden size $H$.

\item For DSS${}_\text{SOFTMAX}$ shown in Table \ref{table:softmax}, the highest accuracy after training with ``Proposed Method'' was 0.8412 at $N=8$.
Notably, this exceeded the accuracy after training with ``HiPPO'' at $N=128$, which was 0.8216, despite having a smaller $N$ while maintaining the same hidden size $H$.
\end{itemize}

In summary, the initial parameters obtained by reducing Pre-Trained DSS of $(H,N)=(16,16)$ for DSS${}_\text{EXP}$ and $(H,N)=(16, 128)$ for DSS${}_\text{SOFTMAX}$ appear to be effective in enhancing accuracy of the trained model compared to the initial parameters by the Skew-HiPPO initialization.
Similar trends are observed in the ListOps task and text retrieval task of LRA, where our method enhanced the accuracy of the trained model.
For detailed results, refer to Appendix \ref{appx:result}.

\subsection{Relationship between accuracy of Pre-Trained models and models after Main Training}

Table\,\ref{table:compare-pre} shows the accuracy of models after Main Training when initialized with different Pre-Trained models.
We obtained Pre-Trained models with DSS of $N=128$ and $N=80$, and utilized the reduced models for improved initialization of Main Training.
The accuracy of the models after main training is in columns ``Proposed Method ($N=128$)'' and ``Proposed Method ($N=80$)''.
Both ``Proposed Method ($N=128$)'' and ``Proposed Method ($N=80$)'' followed the trend observed in Subsection \ref{subsec:result}, where ``Proposed Method'' achieved higher accuracy than ``HiPPO'' for each $N$.

The accuracy of the Pre-Trained models for $N=128$ is 0.8216, which is higher than 0.8098 at $N=80$.
As for models after Main Training, the accuracy of ``Proposed Method ($N=128$)'' surpasses that of ``Proposed Method ($N=80$)'' for each $N$.
This suggests that higher accuracy of the Pre-Trained model leads to higher accuracy of the model obtained through Main Training.

\begin{table}[t]
    \caption{Accuracy of models initialized using different Pre-Trained models (DSS${}_\text{SOFTMAX}$, $H=16$)}
    \label{table:compare-pre}
    \centering
    \begin{tabular}{c c c c}
        \hline
        $N$ & \textbf{HiPPO} & 
        \begin{tabular}{c}
             \textbf{Proposed Method}\\
             ($N=128$) 
        \end{tabular}
        &
        \begin{tabular}{c}
             \textbf{Proposed Method}\\
             ($N=80$) 
        \end{tabular}\\
        \hline \hline
        128 & \textbf{0.8216} & 0.8359 & \\
        80 & 0.8098 & 0.8390 & \textbf{0.8343}\\
        64 & 0.8158 & 0.8382 & 0.8224\\
        32 & 0.8026 & 0.8370 & 0.8255\\
        16 & 0.8190 & 0.8402 & 0.8294\\
        8 & 0.8071 & \textbf{0.8412} & 0.8334\\
        4 & 0.7948 & 0.8377 & 0.8317\\
        \hline
    \end{tabular}
\end{table}

\subsection{Non-triviality of the obtained results}

\begin{figure}[t]
    \centering
    \includegraphics[width=8.5cm]{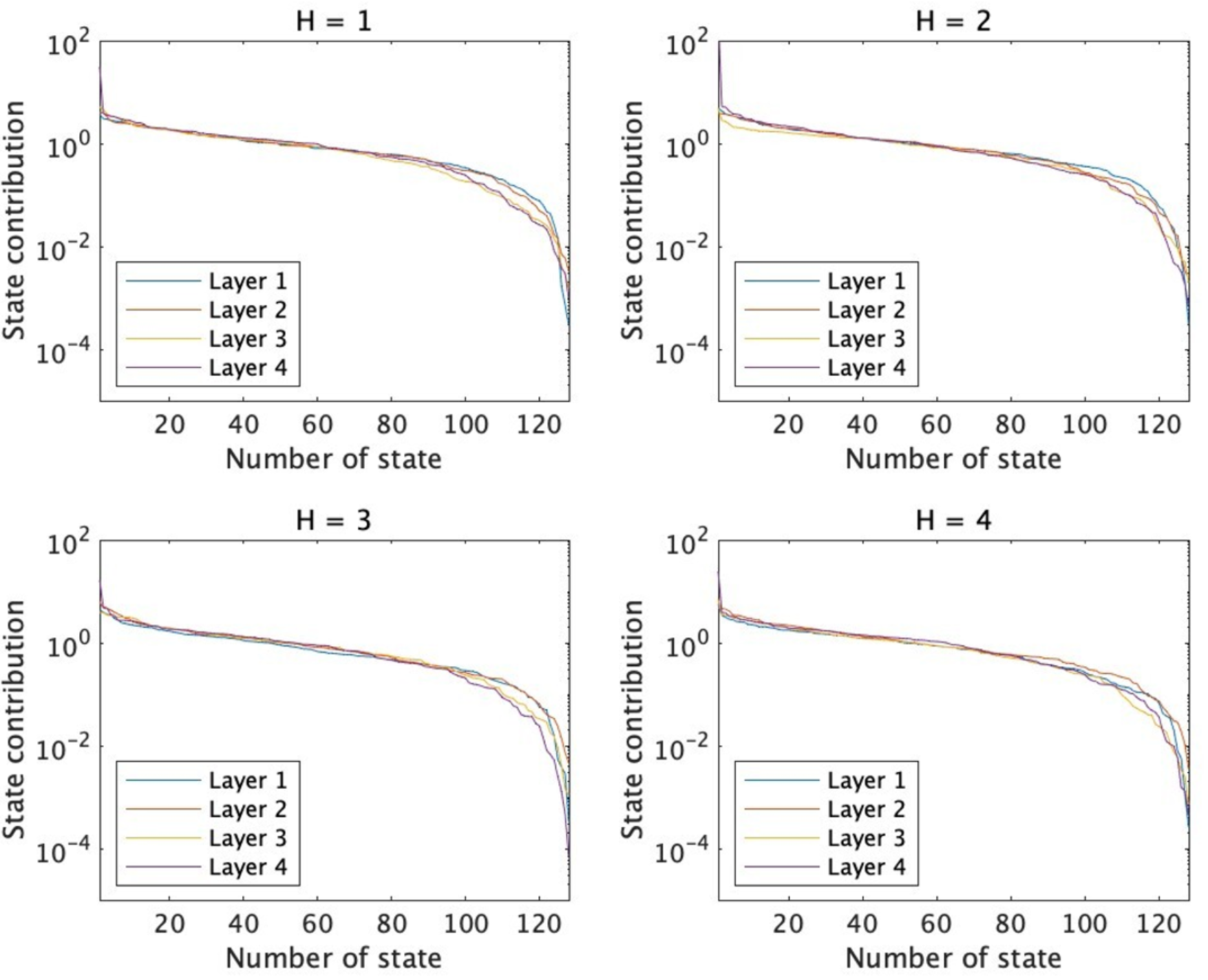}
    \caption{\textcolor{black}{Hankel singular values obtained from each SSM for
    DSS${}_\text{SOFTMAX}$ with $N=128$.
    These SSMs were part of the Pre-Trained model initialized using the Skew-HiPPO with $N=128$, as shown in Table \ref{table:softmax}. Although the hidden size was set to $16$, we only presented the cases for $H=1$, $2$, $3$, and $4$ because the results for $H=5, 6, \ldots, 16$ were almost identical.
    }}
    \label{fig:hsv}
  \end{figure}

\textcolor{black}{
The Hankel singular values illustrated in Fig.\,\ref{fig:hsv} highlight the non-triviality of the results presented in Tables \ref{table:exp} and \ref{table:softmax} from a system-theoretic perspective.
These values were derived from the SSM parameters $(A, B, C)$ of the Pre-Trained model using DSS${}_\text{SOFTMAX}$ with $N=128$.
Specifically, the Hankel singular values were computed for each SSM in every DSS layer. The detailed computational method is described in Appendix \ref{appedix_BT}.}

\textcolor{black}{
As explained in Section \ref{subsec:balred} and Appendix \ref{appedix_BT}, the Hankel singular values can reveal the important directions in the state space from the controllability and observability perspective.
That is, if the Hankel singular values are relatively large, the corresponding directions are relatively important.
Notably, Fig.\,\ref{fig:hsv} shows that almost all directions in the $N=128$ dimensional state space are important, because there are few significantly small Hankel sigular values.
Therefore, reducing the dimensionality of the Pre-Trained model from $N=128$ is expected to significantly deteriorate its performance.
This expectation is consistent with the results shown in Table \ref{table:softmax}. In fact, the accuracy of the Pre-Trained model with $N=128$ was $0.8216$, but after reducing $N$ to $8$, the accuracy dropped to $0.5036$.
Nevertheless, after Main-Training, the accuracy of the reduced model improved to $0.8412$. 
This improvement in accuracy is not predicted by the theoretical analysis of balanced truncation introduced in Appendix \ref{appedix_BT} and is a non-trivial result.}

\section{Conclusion} \label{sec:conclusion}

\textcolor{black}{
We developed a new model compression method specifically for S4 models with DSS layers, using the balanced truncation method \cite{moore1981principal}.
This approach not only reduces the number of parameters but also enhances model performance.
We proposed using the reduced model parameters obtained by the balanced truncation as initial parameters for the main training process.
Our experiments demonstrated that the proposed method achieves superior accuracy on Long Range Arena (LRA) tasks compared to conventionally trained models using the Skew-HiPPO initialization, even with fewer parameters.}
Moreover, we observed a positive correlation between the accuracy of Pre-Trained models and their accuracy after Main Training.

\textcolor{black}{
The primary limitation of this study lies in the scope of tasks and datasets used for evaluation. While the LRA tasks provide a robust benchmark for long-range dependency modeling, further validation on diverse datasets and real-world applications is necessary. Additionally, the underlying principles of the proposed method remain unclear, which limits the understanding of why this approach is effective.
}

\textcolor{black}{
The following are interesting future directions:
}
\begin{itemize}
    \item 
\textcolor{black}{
    Future research should investigate the underlying principles of the proposed method, aiming to enhance the development of more effective training methods for deep learning models with DSS layers.}

\item
\textcolor{black}{
Reference \cite{DBLP:journals/corr/HanMD15}
has shown that combining various model compression methods can yield better results. Investigating whether combining our proposed model compression method based on the balanced truncation with other compression techniques can improve performance is an interesting and promising direction for future work.
}

    \item 
\textcolor{black}{
    Expanding the scope of evaluation to include real-time deployment scenarios in EI applications will provide more comprehensive insights into the method's practical viability.
    This can help demonstrate how the reduced models can be effectively used in resource-constrained environments.}
    
    \item 
\textcolor{black}{
    As an extension of our study, applying the proposed compression techniques to Physics-Informed Neural Networks (PINNs) can be explored \cite{donnelly2024physics, raissi2019physics}. This could help to improve the efficiency and performance of PINNs in modeling physical systems with limited computational resources.}
\end{itemize}

 \section*{Acknowledgment}
 This work was supported by the Japan Society for the Promotion of Science KAKENHI under Grant 23H03680.

\appendices

\section{Details of the balanced truncation method}\label{appedix_BT}

\textcolor{black}{
As mentioned in Section \ref{subsec:balred}, the eigenvectors of the controllability and observability Gramians $P$ and $Q$ of SSM \eqref{eq:c-SSM} provide important directions in the state space $\mathbb{C}^N$ from the perspectives of controllability and observability.
Thus, we can adopt an approach that reduces dimensions along directions that are not significant.
However, the eigenvectors do not coincide in general.
This means that, in general, it is impossible to uniquely determine the directions to be ignored based solely on the information from the original controllability Gramian $P$ and observability Gramian $Q$.}

\textcolor{black}{
To overcome this problem, we apply a coordinate transformation $\bar{x}(t)=Tx(t)$ to SSM \eqref{eq:c-SSM} to obtain new SSM \eqref{SSM_change}.
Then, the corresponding controllability and observability Gramians of SSM \eqref{SSM_change} become $TPT^*$ and $(T^{-1})^*QT^{-1}$, respectively.
Thus, if we can find $T\in \mathbb{C}^{N\times N}$ satisfying
\begin{align}
TPT^*=(T^{-1})^*QT^{-1}, \label{eq_PQ_henkan0}    
\end{align}
 the controllability and observability Gramians of the transformed SSM \eqref{SSM_change} will coincide, even if the original Gramians of SSM \eqref{eq:c-SSM} do not.}

\textcolor{black}{
To find $T$ satisfying \eqref{eq_PQ_henkan0}, 
we perform the eigenvalue decomposition of the symmetric positive definite matrix $P^{1/2}QP^{1/2}$ to obtain
\begin{align}
    P^{1/2}QP^{1/2} = U\Lambda U^*, \label{PQP_eig}
\end{align}
where $P^{1/2}$ is the square root matrix of $P$, $U$ is a unitary matrix, and $\Lambda=\mathrm{diag}(\lambda_1,\ldots, \lambda_N)$ is a diagonal matrix with positive diagonal elements satisfying $\lambda_1\geq \cdots \geq \lambda_N$.
Defining 
\begin{align}
T:=\Lambda^{1/4}U^*P^{-1/2},  \label{eq_T}
\end{align}
 we get
\begin{align}
    TPT^* = (T^{-1})^*QT^{-1} = \Lambda^{1/2}=\mathrm{diag}(\sigma_1,\ldots, \sigma_N). \label{eq_PQ_henkan}
\end{align}
We call $\sigma_i=\sqrt{\lambda_i}$ $(i=1,\ldots, N)$ the Hankel singular values of SSM \eqref{eq:c-SSM}.}

\textcolor{black}{
Thus, the balanced truncation method consists of the following procedure:}
\begin{enumerate}
    \item \textcolor{black}{Compute the controllability Gramian $P$ and the observability Gramian $Q$ of SSM \eqref{eq:c-SSM} by solving the Lyapunov equations \eqref{Lyap_P} and \eqref{Lyap_Q}, respectively.}

    \item \textcolor{black}{Compute the square root matrix $P^{1/2}$.}

    \item \textcolor{black}{Perform the eigenvalue decomposition using \eqref{PQP_eig}.}

    \item \textcolor{black}{Determine the transformation matrix $T$ using \eqref{eq_T}.}

    \item \textcolor{black}{Compute \eqref{A_trans}, \eqref{B_trans}, and \eqref{C_trans} using $T$, and define $(A_r, B_r, C_r)$ of reduced SSM \eqref{eq:r-dss} as \eqref{matrix_reduced}.}
\end{enumerate}

\textcolor{black}{
The reduced SSM \eqref{eq:r-dss}
is preferable when it closely approximates the original large-scale SSM \eqref{eq:c-SSM} in terms of the $H^{\infty}$ norm of the difference in their transfer functions.
In fact, the transfer functions $G$ and $G_r$ of original SSM \eqref{eq:c-SSM} and  reduced SSM \eqref{eq:r-dss} are defined by
\begin{align}
    G(s):= C(sI_N-A)^{-1}B,\quad G_r(s):=C_r(sI_r-A_r)^{-1}B_r,
\end{align}
respectively. The energy of the difference between original SSM \eqref{eq:c-SSM} output $y$ and reduced SSM \eqref{eq:r-dss} output $y_r$ can be evaluated by
\begin{align}
    \int_0^\infty \|y(t)-y_r(t)\|^2 {\rm d}t \leq \|G-G_r\|_{H^\infty}^2     \int_0^\infty \|u(t)\|^2 {\rm d}t,
\end{align}
where $\|\cdot\|_{H^\infty}$ denotes the $H^\infty$ norm. That is, if the input energy $\int_0^\infty \|u(t)\|^2 {\rm d}t$ and
$\|G-G_r\|_{H^\infty}$
are sufficiently small, the output error energy $\int_0^\infty \|y(t)-y_r(t)\|^2 {\rm d}t$ is also sufficiently small.
Moreover, $\|G-G_r\|_{H^\infty}^2$ when using the balanced truncation method is bounded by
\begin{align}
    \sigma_{r+1} \leq \|G-G_r\|_{H^{\infty}} \leq 2(\sigma_{r+1}+\cdots +\sigma_N),
\end{align}
assuming that $\sigma_{r+1}>\cdots >\sigma_N$.
Thus, if the Hankel singular values $\sigma_{r+1}, \ldots, \sigma_N$ are small, then $\|G-G_r\|_{H^{\infty}}$ will also be small.
The proofs for the above claims can be found in \cite[Chapter 4]{dullerud2000course}.
}

\section{Details of the deep learning model} \label{appx:whole-model}

\begin{figure}[t]
    \centering
    \includegraphics[width=8.5cm]{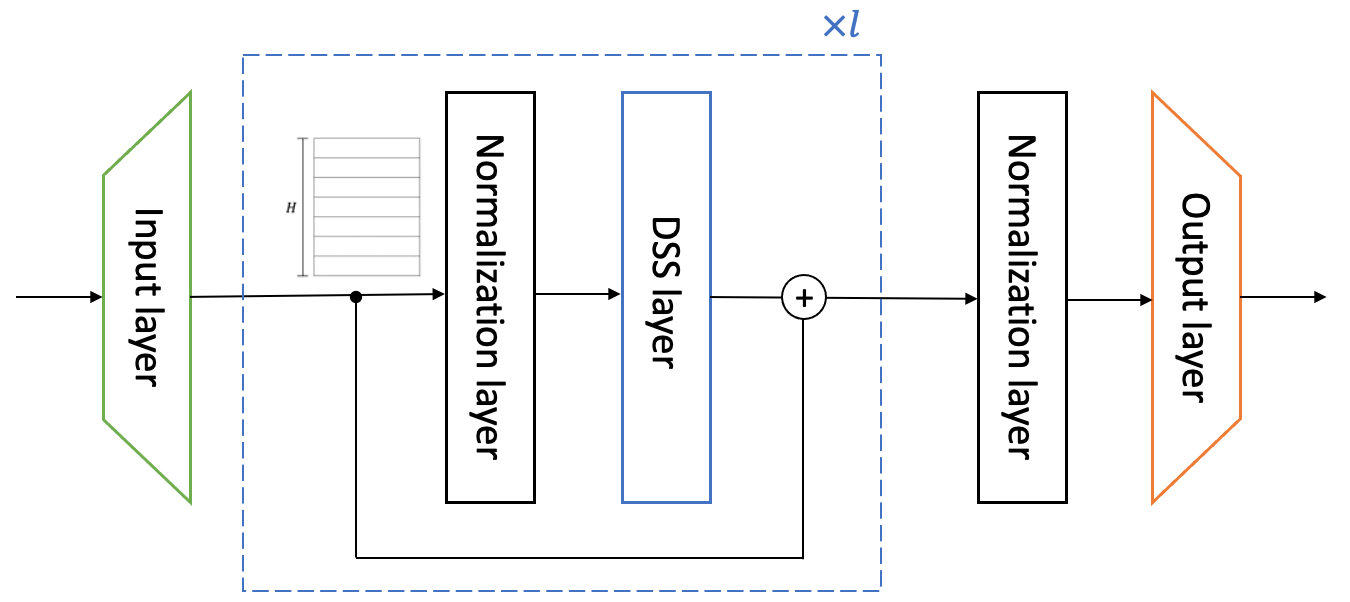}
    \caption{Deep learning model with DSS layers (Prenorm). \textcolor{black}{The normalization layer is placed before the DSS layer and output layer.}}
    \label{fig:prenorm-model}
  \end{figure}

\begin{figure}[t]
    \centering
    \includegraphics[width=8.3cm]{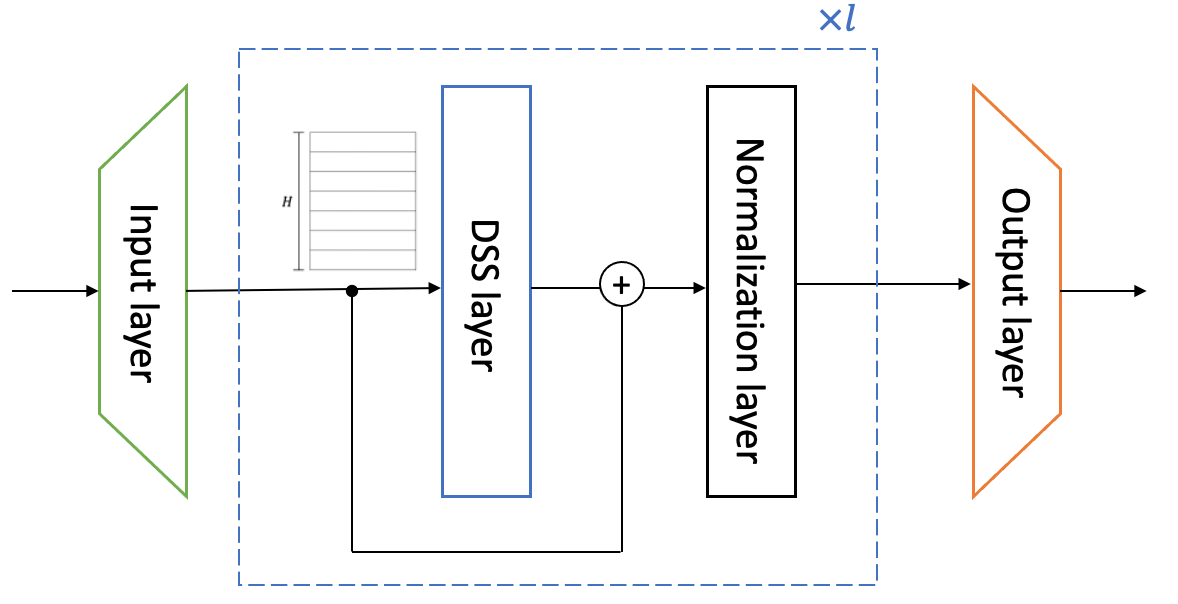}
    \caption{Deep learning model with DSS layers (Postnorm). \textcolor{black}{The normalization layer is placed after the residual connection.}}
    \label{fig:postnorm-model}
\end{figure}

In the deep learning model employed in this study, residual connections \cite{he2016deep} and normalization layers are positioned before and after the DSS layer.
In residual connections, a path bypassing one or more layers is created, as illustrated in Fig.\,\ref{fig:prenorm-model} and Fig.\,\ref{fig:postnorm-model}, and the output of the bypassed layer is added to it.
The normalization layer can be placed before the DSS layer (Prenorm, Fig.\,\ref{fig:prenorm-model}) or after the residual connection (Postnorm, Fig.\,\ref{fig:postnorm-model}).
In the case of prenorm, a normalization layer is also placed before the output layer.
Normalization layers such as batch normalization \cite{ioffe2015batch} or layer normalization \cite{ba2016layer} are used, which contributes to the stability and acceleration of training.
Additionally, residual connections prevent the gradient vanishing and exploding problems.


\section{Other results} \label{appx:result}

In addition to the text classification task, explained in Section \ref{sec:experiments},
we confirmed that our proposed method improves performance in the ListOps task and the text retrieval task of LRA \cite{tay2021long}, as shown in Table\,\ref{table:listops} and Table\,\ref{table:aan}, respectively.
\textcolor{black}{Here, the number of DSS layers is $6$ and the hidden size $H$ is $16$.
}
In the ListOps task, a numerical expression structured with operators MAX, MEAN, MEDIAN, SUM\_MOD and parentheses is the input, and its value is the output.
For instance,
\begin{align}
    \begin{split}
        & \mathtt{INPUT:[~MAX~1~2~MIN~[~3~4~]~MEDIAN~[~1~5~9~]~]} \\
        & \mathtt{OUTPUT:5}        
    \end{split}
\end{align}
The maximum length of input is 2000, and the output values range from 0 to 9.
In the text retrieval task, we estimate the similarity between two papers and determine if there is a citation link.
The length of each paper is 4000, and the total input length is 8000.

\begin{table}[t]
    \caption{Accuracy of models through various training methods (Listops, DSS${}_\text{EXP}$, $H=16$)}
    \label{table:listops}
    \centering
    \begin{tabular}{c c c c c}
        \hline
        $N$ &  & \textbf{HiPPO} & \textbf{Random} & \textbf{Proposed Method} \\
        \hline
        \hline
        \multirow{2}{*}{64}  & before & 0.0715 & 0.0810 & 0.5180 \\
                             & after & \textbf{0.5180} & 0.4020 & 0.5300 \\
        \hline
        \multirow{2}{*}{32}  & before & 0.1705 & 0.1195 & 0.3610 \\
                             & after & 0.4920 & 0.4035 & 0.5205 \\
        \hline
        \multirow{2}{*}{16}  & before & 0.0760 & 0.1780 & 0.1290 \\
                             & after & 0.4745 & 0.4001 & \textbf{0.5390} \\
        \hline
        \multirow{2}{*}{8}   & before & 0.0715 & 0.0960 & 0.1825 \\
                             & after & 0.4025 & 0.4300 & 0.5250 \\
        \hline
        \multirow{2}{*}{4}   & before & 0.0825 & 0.1210 & 0.1725 \\
                             & after & 0.4250 & 0.4440 & 0.5175 \\
        \hline
    \end{tabular}
\end{table}

\begin{table}[t]
    \caption{Accuracy of models through various training methods (Text retrieval, DSS${}_\text{EXP}$, $H=16$)}
    \label{table:aan}
    \centering
    \begin{tabular}{c c c c c}
        \hline
        $N$ &  & \textbf{HiPPO} & \textbf{Random} & \textbf{Proposed Method} \\
        \hline
        \hline
        \multirow{2}{*}{64}  & before & 0.4943 & 0.5068 & 0.8189 \\
                             & after & 0.8189 & 0.7830 & 0.8345 \\
        \hline
        \multirow{2}{*}{32}  & before & 0.4937 & 0.4976 & 0.5807 \\
                             & after & \textbf{0.8217} & 0.7812 & 0.8302 \\
        \hline
        \multirow{2}{*}{16}  & before & 0.5055 & 0.4932 & 0.5399 \\
                             & after & 0.8071 & 0.7907 & 0.8251 \\
        \hline
        \multirow{2}{*}{8}   & before & 0.4939 & 0.4939 & 0.5064 \\
                             & after & 0.8212 & 0.7944 & 0.8270 \\
        \hline
        \multirow{2}{*}{4}   & before & 0.4939 & 0.4939 & 0.5062 \\
                             & after & 0.8136 & 0.7893 & \textbf{0.8313} \\
        \hline
    \end{tabular}
\end{table}

For DSS${}_\text{SOFTMAX}$, the left column ``HiPPO'' using the Skew-HiPPO initialization achieved higher accuracy after training compared to the middle column ``Random'' using randomly sampled initial values.
For DSS${}_\text{EXP}$, the same trend was observed for almost all $N$ as shown in Table\,\ref{table:listops} and Table\,\ref{table:aan}.

In the ``Proposed Method'' column on the right, Main Training was initialized using a reduced model obtained from Pre-Trained models of $(H, N)=(16, 64)$.
The accuracy of models before Main Training with ``Proposed Method'' was comparable to that of ``Random'' and ``HiPPO'' for each $N$ excluding $N=64$.
However, after the training, the accuracy of ``Proposed Method'' exceeded that of ``HiPPO'' for each $N$.

The following points are particularly noteworthy.
\begin{itemize}
    \item For ListOps task, the highest accuracy after Main Training with ``Proposed Method'' was 0.5390 at $N=16$.
Notably, this exceeded the accuracy after training with ``HiPPO'' at $N=64$, which was 0.5180, despite the smaller $N$ and the same hidden size $H$.

\item For text retrieval task, the accuracy after Main Training with ``Proposed Method'' was 0.8313 at $N=4$, which exceeded the accuracy after training with ``HiPPO'' at $N=64$ and $N=32$, despite the smaller $N$ and the same hidden size $H$.
\end{itemize}

Consequently, the initial parameters obtained by reducing Pre-Trained DSS of $(H, N)=(16, 64)$ appear to be effective in enhancing accuracy of the trained model compared to the initial parameters by the Skew-HiPPO initialization.





\bibliographystyle{IEEEtran}
\bibliography{main.bib}




%




\end{document}